\documentclass{article}

    \PassOptionsToPackage{numbers, compress}{natbib}

\usepackage[final]{neurips_2022}

\usepackage[utf8]{inputenc} 
\usepackage[T1]{fontenc}    
\usepackage{hyperref}       
\usepackage{url}            
\usepackage{booktabs}       
\usepackage{amsfonts}       
\usepackage{nicefrac}       
\usepackage{xcolor}         
\usepackage{amsmath}
\usepackage{amsthm}
\usepackage{graphicx}
\usepackage{subcaption}
\usepackage{wrapfig}
\usepackage{dsfont}
\usepackage{multirow}
\usepackage{cancel}
\usepackage{pgfplots}
\usepackage{threeparttable}

\usepackage{tikz}
\usetikzlibrary{positioning,fit,calc}
\usepgfplotslibrary{groupplots}
\pgfplotsset{width=7.5cm,compat=1.15, tick label style={font=\small}}

\usepackage{xspace}

\makeatletter
\DeclareRobustCommand\onedot{\futurelet\@let@token\@onedot}
\def\@onedot{\ifx\@let@token.\else.\null\fi\xspace}

\makeatother

\newcommand{\Expect}{\mathbb{E}}
\newcommand{\Prob}{\mathbb{P}}
\newcommand{\pout}{p_{\text{out}}}
\newcommand{\poutp}{p_{\text{out}}^{\text{proxy}}}
\newcommand{\pin}{p_{\text{in}}}



\definecolor{cornellred}{RGB}{179,27,27}
\definecolor{cornellblue}{RGB}{0,104,172}
\definecolor{cornellgreen}{RGB}{110,180,63}
\definecolor{cornellgrey}{RGB}{85,86,90}
\definecolor{mypurple}{RGB}{116,79,198}

\title{Falsehoods that ML researchers believe about OOD detection}

\author{%
  Andi Zhang \\
  Computer Laboratory\\
  University of Cambridge\\
  \texttt{az381@cam.ac.uk} \\
  \And
  Damon Wischik \\
  Computer Laboratory\\
  University of Cambridge \\
  \texttt{djw1005@cam.ac.uk} \\
}

\begin{document}

\maketitle

\begin{abstract}
An intuitive way to detect out-of-distribution (OOD) data is via
the density function of a fitted probabilistic generative model:
points with low density may be classed as OOD. But this approach has been found
to fail, in deep learning settings.
In this paper, we list some falsehoods that machine learning researchers believe about density-based OOD detection. 
Many recent works have proposed likelihood-ratio-based methods to `fix' the
problem. 
We propose a framework, the \emph{OOD proxy framework}, to unify these methods, and we argue that likelihood ratio is a principled method
for OOD detection and not a mere `fix'.
Finally, we discuss the relationship between domain discrimination and semantics.
\end{abstract}

\section{Introduction}

We might expect that a neural network should produce reliable outputs when it is 
presented with data similar to that used in training, and that its outputs might
be prone to error when it is presented with substantially different data. It is clearly
desirable for a neural network to be able to detect the latter case. This is called
the out-of-distribution (OOD) detection problem~\citep{hendrycks2016baseline}.

A naive approach to OOD detection is as follows: First, train
a density model $p(x)$ to approximate the true distribution from which the training 
dataset is assumed to be drawn. If $p(x)$ is small at some particular novel input $x$,
it indicates that there is little training data in the region around $x$, and that the model 
should therefore be unconfident.

This naive approach leads to a paradoxical result, as elegantly shown by
\citet{nalisnick2018deep}. They found that if they train a generative model
to learn the density $p(x)$ on CIFAR10, and then evaluate this trained $p(x)$ on two test
sets, one from CIFAR10 and one from SVHN, then the test CIFAR10 scores
are \emph{lower} than those for SVHN. They found this paradoxical result in several
other examples, and it is easy to replicate.

In this paper, we will argue that this result is not in fact paradoxical: that in fact
the naive approach to OOD detection is based on several falsehoods, falsehoods which are
readily demonstrated using basic probability and statistics. These falsehoods are
\begin{itemize}\setlength{\itemsep}{0em}
\item that $p(x)$ should be lower on OOD data;
\item that the paradoxical result arises from some deep-learning dark magic;
\item that $p(x)$ is suitable for comparing two distributions;
\item that low $p(x)$ indicates lack of samples.
\end{itemize}
We will also argue that some successful approaches to OOD detection in the literature
(starting with \citet{bishop1994novelty}) are based on the likelihood ratio between
\emph{two} datasets rather than the density for a single dataset, and that the substantive
differences are to do with how this second dataset is constructed.

\section{Falsehoods}
\label{sec:falsehoods}

Here is a simple example to illustrate the problems with using $p(x)$ for OOD detection.
Suppose the training dataset is drawn from $N(0,1)$, and that the training procedure
has correctly learned the density $p(x)=\mathcal{N}(x;0,1)$. 
Now consider an OOD dataset
drawn from $N(0,\varepsilon^2)$ for some small $\varepsilon$. Then the expected log likelihoods
are
\begin{align*}
    \Expect \log p(X) = 
    \frac{1}{2}\log 2\pi - \begin{cases}
    1/2 &\text{for in-distribution i.e. } X\sim N(0,1)\\
    \varepsilon^2/2 & \text{for OOD i.e. } X\sim N(0,\varepsilon^2).
    \end{cases}
\end{align*}
We see that $\log p(X)$ is larger for out-of-distribution data. This isn't a paradox, it's 
expected behaviour! And it arises from basic probability, not from mysterious properties of 
deep generative modelling.

This phenomenon does not seem to be limited to toy examples. \citet{nalisnick2018deep} believe
it can hold in real-world datasets, and explains their finding that $p(x)$ fitted to CIFAR-10 data is no good
for detecting OOD datapoints from SVHN:
`Our conclusion is that SVHN simply ``sits inside of'' CIFAR10---roughly same mean, 
smaller variance---resulting in its higher likelihood.'

\paragraph{Outlier detection \emph{v.} OOD detection.}
It is entirely reasonable to use $p(x)$ to test whether a datapoint is an outlier. This is 
indeed the cornerstone of frequentist statistics---we reject the null hypothesis when the test statistic
shows that the observed data is unlikely. It's a reasonable basis for anomaly detection, to say
`if $p(x)$ is low then label $x$ as an outlier.''~\citep{ruff2021unifying}

But OOD detection isn't the same thing as outlier detection.
In the two examples above, we implicitly used the phrase `OOD detection' to mean ``drawn from a specified other distribution'', and this other
distribution happened to include non-outlier points.

In conclusion, $p(x)$ is not suitable for comparing two distributions. In section~\ref{sec:lr}, we will discuss how to compare two distributions properly.




\paragraph{Lack of samples?}
Why was the result of \citet{nalisnick2018deep} surprising? The intuition is something like this:
the training dataset (CIFAR10) has no samples that look anything like the OOD dataset (SVHN),
therefore we expect $p(x)$ to be low on those OOD datapoints.

But this intuition breaks down in high dimensions. A well-known result says that, with high probability, samples
drawn a from high-dimensional Gaussian lie in a thin annulus --- ``Gaussian distributions are soap bubbles''~\citep{blum2020foundations}.
The pdf is always highest at the origin, and yet we are very unlikely to see any sample points in a ball around the origin!~\citep{nalisnick2019detecting}

In other words, ``lack of samples'' should not be confused with ``low pdf''.



\section{Likelihood ratio}
\label{sec:lr}

\citet{bishop1994novelty} pointed out that OOD detection can be thought of as model selection between the in-distribution $p_{\text{in}}$ and an out-of-distribution $p_{\text{out}}$. 

In frequentist terminology, given an observation $x$, consider the null hypothesis $H_0$ that $x$ was drawn from $p_{\text{in}}$, and the alternative
hypothesis $H_1$ that $x$ was drawn from $p_{\text{out}}$. By the Neyman-Pearson lemma~\cite{neyman1933ix}, when fixing type-I error $P(\text{reject }H_0 | H_0\text{ is true})$, the test with the smallest type-II error $P(\text{accept }H_0 | H_0 \text{ is false})$ is the likelihood ratio test. 
This implies that using likelihood ratio as a test score will optimise the area under the receiver operating characteristic (AUROC), which is a popular OOD detection baseline suggested by \citet{hendrycks2016baseline}.

A similar result holds under the Bayesian perspective. Let $C\sim\operatorname{Bin}(1,\alpha)$, 
and let $X\sim p_{\text{in}}$ if $C=0$ and $X\sim p_{\text{out}}$ if $C=1$.
Given an observed value $x$,  
\begin{align*}
    \Prob(C=1 | x) = \frac{p(x|C=1)\,\Prob(C=1)}{p(x|C=1)\,\Prob(C=1) + p(x|C=0)\,\Prob(C=0)} = \frac{1}{1+(1-\alpha)/(\alpha\, \textit{LR})}
\end{align*}
where $\textit{LR} = p_{\text{out}}(x)/p_{\text{in}}(x)$ is the likelihood ratio. Since $\Prob(C=1|x)$ is an increasing function of $\textit{LR}$,
we might set a threshold $\theta$ and decide $C=1$ if $\textit{LR}>\theta$.

We have shown that the likelihood ratio is an optimal choice from both frequentist and Bayesian perspectives. However, it is hard to obtain $p_\text{out}$. In the next section, we introduce some practical works that propose proxies for $p_{\text{out}}$.

\section{OOD proxies}
\label{sec:proxy}

In practice, we typically do not have an explicit $\pout$ distribution. Several recent works on OOD detection can however
be thought of as using a likelihood ratio test based on a \emph{proxy} distribution for $\pout$. Formally, 
we can propose an OOD proxy $p_{\text{out}}^{\text{proxy}}$, 
and use the likelihood ratio $p_{\text{in}}/p_{\text{out}}^{\text{proxy}}$ as our OOD criterion. 


\paragraph{Constant.}
\citet{bishop1994novelty} suggested we take $p_{\text{out}}^{\text{proxy}}$ to be a constant. This expresses the intuitive idea that $p_{\text{out}}$ should spread
out widely in a large area. Given $x$, the likelihood ratio between $p_{\text{in}}(x)$ and a constant is proportional to $p_{\text{in}}(x)$, which is identical to the criterion $p(x)$ used by \citet{nalisnick2018deep} if we ignore the scale of the threshold. 
They reported that this choice of $\pout^{\text{proxy}}$ leads to poor performance, as measured by AUROC,
in deep learning examples.

\paragraph{Auxiliary OOD datasets.}

It is natural to construct $p_{\text{out}}^{\text{proxy}}$ by some real out-of-distribution data. \citet{hendrycks2018deep} suggested that introducing an auxiliary OOD data\footnote{To have a fair comparison in the benchmark introduced by \citet{hendrycks2016baseline}, the auxiliary OOD dataset does not have any intersection with the test OOD dataset.} (e.g.~80 Million Tiny Images~\citep{deng2009imagenet}) will increase the anomaly detection performance. Here, the auxiliary OOD datasets play a role of the $p_\text{out}^{\text{proxy}}$. \citet{hendrycks2018deep} did not use the likelihood ratio as the criterion for OOD detection, they proposed to fine-tune the generative model by the loss
$$\max\{0, C -\log p(x_\text{in}) + \log p(x_\text{out})\}$$
where $C$ is a the number of the pixels of the image, $x_\text{in}$ is the in-distribution data and $x_\text{out}$ is the out-of-distribution data. Then they keep using the likelihood $p(x)$ to detect OOD. Following their OOD proxy, \citet{schirrmeister2020understanding} proposed a criterion using likelihood ratio between in-distribution $p_{\text{in}}$ and general image distribution $p_{\text{g}}$, where $p_{\text{g}}$ is trained by the aforementioned auxiliary OOD dataset, i.e.~the $p_\text{out}^{\text{proxy}}$. Furthermore, \citet{zhang2022out} suggested that the likelihood ratio could be estimated by a binary classifier. 

\paragraph{Background statistics.}

\citet{ren2019likelihood} observed that ``the background of images confounds the likelihood of the generative models'', and propose a method
for OOD detection based on eliminating the effect of background. Assume that background and semantic components are generated independently,
i.e. $p(x)=p(x_S)\, p(x_B)$ where $x_S$ stands for semantics and $x_B$ stands for background.
Suppose we know this factorization for the in-distribution data, as well as for proxy OOD data which has been obtained by perturbing the in-distribution
data in such a way as to preserve the background and lose the semantics.
They propose using the likelihood ratio $\pin(x_S)/\poutp(x_S)$ for OOD detection.

In practice, it's hard to see how we can learn this factorization into semantics and background. They propose instead
that $\pin(x_B)\approx \poutp(x_B)$, since we perturbed the data so as to preserve the background. Then their likelihood ratio becomes
\[
\textit{LR}(x) 
= \frac{\pin(x_S)}{\poutp(x_S)} 
\approx \frac{\pin(x_S)\, \pin(x_B)}{\poutp(x_S)\, \poutp(x_B)}
= \frac{\pin(x)}{\poutp(x)}
\]
which is exactly our general-purpose likelihood ratio criterion.

\paragraph{Input complexity.} 

\citet{serra2019input} observed that realistic in-distribution images typically have higher complexity, and that higher-complexity
images typically have low $p(x)$, and suggest this is why using $p(x)$ is not good for OOD detection. They propose compensating
for this by using a score $S(x)=\log_2 \pin(x)+L(x)$ where $L$ is a measure of image complexity: the number of bits when $x$ is compressed by
a universal compressor. They point out that this is effectively a likelihood ratio test, using an OOD proxy distribution
$\poutp(x)\propto 2^{-L(x)}$. Similar to Ren et al., they interpret $\poutp$ as describing the background without specific semantics.


\paragraph{Local features.}
\citet{zhang2021out} proposed detecting OOD by using 
local models, i.e. models constrained to capture only limited perceptual fields of the image. 
They observed that the local models and full models assign similar likelihoods to OOD data, and infer that
the local features are shared between in-distribution and OOD datasets while non-local features are not.
They assume that the full model admits a decomposition $\pin(x) \propto \pin^{\text{local}}(x)\,\pin^{\text{nonlocal}}(x)$,
and propose that $\pin^{\text{nonlocal}}$ should be used for detecting OOD data. This can be written as
\[
\pin^{\text{nonlocal}}(x) \propto \frac{\pin(x)}{\pin^{\text{local}}(x)}
\]
which is our general-purpose likelihood ratio criterion, using the local model trained on in-distribution data as the proxy OOD distribution.

\paragraph{Label-based.} 
Suppose we're trying to detect OOD inputs to a classifier which we've trained on an
dataset of ($x$,label) pairs. \citet{hendrycks2016baseline} suggested
using the predicted labels for OOD detection: for example, if $y(x)$ is the vector of
class probabilities predicted for input $x$, they suggest labelling $x$ as OOD if
the entropy $H\bigl(y(x)\bigr)$ is above a threshold. This idea has been taken
on by others 
\citep{lee2018simple, liang2017enhancing, malinin2018predictive, sastry2020detecting, sensoy2018evidential}.
We can interpret the entropy-based detector
as a likelihood ratio test, comparing $\pin$ to $\poutp$ defined by
\[
\poutp(x) \propto e^{H(y(x))}\, \pin(x).
\]
It's interesting to speculate what this proxy distribution might look like;
we are not aware of any work on this.


\section{Discussion}

\paragraph{Semantics \emph{v.} domain distinction.}

The works we have discussed~\citep{ren2019likelihood, schirrmeister2020understanding, serra2019input, zhang2021out, zhang2022out}
include interpretations in the language of semantics. 
Indeed, the benchmark proposed by \citet{hendrycks2016baseline} is semantic: 
``We can see that SVHN is semantically different to CIFAR10, so SVHN
should be considered OOD.'' But it's hard to define `semantics' rigorously, and so semantic-based OOD detection can seem \emph{ad hoc}. 
In our opinion, it's simpler to
treat OOD detection as just a problem of detecting domains ($\pin$ versus $\poutp$), and this leads directly to the very clean
answer ``use likelihood ratio'' discussed in section~\ref{sec:lr}. In effect, what we propose can be thought of as
\emph{defining} semantics in terms of domains: the semantics of $\pin$ are those features that are absent in $\poutp$.

One case where there is a somewhat clearer understanding of semantics is with
labelled training data: the labels surely capture \emph{some sort} of useful semantics. 
Label-based semantics can be linked to domain distinction, as shown by our
label-based OOD proxy described above.


\paragraph{Likelihood-ratio is not a hack.}
Most of the works introduced in section~\ref{sec:proxy} use `failure' or some similar words to describe the phenomenon reported by \citet{nalisnick2018deep}. They proposed solutions or patches based on background statistics, local features, or data complexity to ``fix the issue''; and all of them have a final form in likelihood ratio. 
According to \citet{bishop1994novelty}, and as we discussed in section~\ref{sec:lr}, density-based OOD detection is a special case of likelihood-ratio-based OOD detection. Hence, we emphasise that likelihood ratio is not a hack to fix density-based detection, it is a principled way to detect OOD.


\paragraph{Generalisation of OOD proxies.}
According to section~\ref{sec:proxy}, it is important to design a proper OOD proxy such that the model is able to distinguish the in-distribution test set $\mathcal{D}_{\text{in}}^\text{test}$ from many different OOD test sets $\mathcal{D}_{\text{out1}}^\text{test}, \mathcal{D}_{\text{out2}}^\text{test},\dots$  In other words, we want an OOD proxy that can distinguish the in-distribution domain from other domains. We call this the generalisation of OOD proxy. \citet{hendrycks2018deep} indicated that using real auxiliary data (e.g.~80 Million Tiny Images~\citep{deng2009imagenet}) as the OOD proxy has a better performance than using the augmented in-distribution data. We believe this is because the real auxiliary data is more similar to the OOD data or has a large intersection with the domains of the OOD test datasets. Investigating the generalisation of OOD proxies is an open question, which we leave to future work.

\clearpage
\newpage

\bibliographystyle{abbrvnat}
\bibliography{ref}


\end{document}